\pdfoutput=1
\documentclass[11pt]{article}

\usepackage[final]{acl}

\newcommand{\phish}{PHISH\includegraphics[height=.75em]{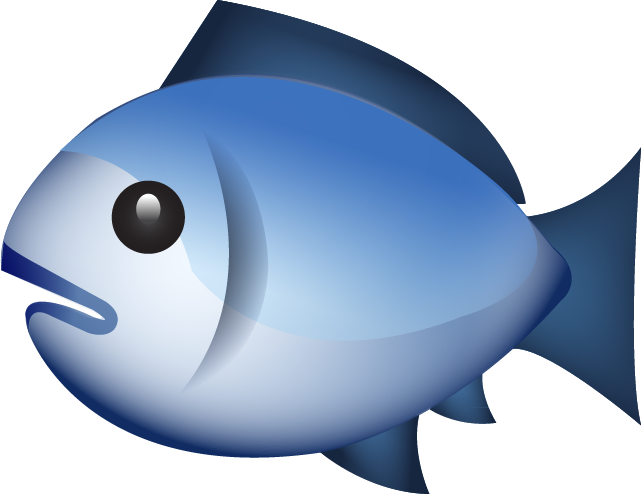}} 
\newcommand{\mesh}{MESH\includegraphics[height=.75em]{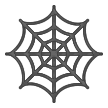}}
\newcommand{\dirmesh}{\textsubscript{dir-MESH}} %\includegraphics[height=.75em]{images/mesh2.png}}}
\newcommand{\seqmesh}{\textsubscript{seq-MESH}} %\includegraphics[height=.75em]{images/mesh2.png}} }

\usepackage{tipa}
\usepackage{booktabs}
\usepackage{times}
\usepackage{latexsym}
\usepackage{kotex}
\usepackage{algorithm}
\usepackage{algpseudocode}
\usepackage{amsmath}
\usepackage{amssymb}
\usepackage{tipa}
\DeclareTextFontCommand\textphonetic{\ipafont}
\usepackage{multirow}
\usepackage{mathtools}
\usepackage{subfig}

\usepackage[T1]{fontenc}
\usepackage[utf8]{inputenc}

\usepackage{microtype}

\usepackage{inconsolata}

\usepackage{graphicx}
\title{\phish\ in \mesh: Korean Adversarial Phonetic Substitution and Phonetic-Semantic Feature Integration Defense}
\author{Byungjun Kim, Minju Kim, Hyeonchu Park, and Bugeun Kim
\\
        Department of Artificial Intelligence, Chung-Ang University, Republic of Korea \\
        \{k36769, minjunim, phchu0429, bgnkim\}@cau.ac.kr}
\begin{document}
\maketitle
\begin{abstract}
% \textit{Warning: this paper contains expressions that
% may offend the readers.}
% \\~\\
As malicious users increasingly employ phonetic substitution to evade hate speech detection, researchers have investigated such strategies. However, two key challenges remain. First, existing studies have overlooked the Korean language, despite its vulnerability to phonetic perturbations due to its phonographic nature. Second, prior work has primarily focused on constructing datasets rather than developing architectural defenses.
To address these challenges, we propose (1) PHonetic-Informed Substitution for Hangul (\phish) that exploits the phonological characteristics of the Korean writing system, and (2) Mixed Encoding of Semantic-pHonetic features (\mesh) that enhances the detector's robustness by incorporating phonetic information at the architectural level.
Our experimental results demonstrate the effectiveness of our proposed methods on both perturbed and unperturbed datasets, suggesting that they not only improve detection performance but also reflect realistic adversarial behaviors employed by malicious users.

% As malicious users often perturb their hateful texts using phonetic substitution, researchers have investigated such strategies. However, we believe that there are two challenges remain. First, existing studies did not deal with the Korean language even its photographic characteristics is vulnerable to such perturbation. Second, prior works focused on constructing datasets to against such attacks, rather architectural methods. To tackle these challenges, we propose a (1) phonetic adversarial substitution attack, PASA, which leverages characteristics of the Korean language, and (2) cumulative or integrative semantic-phonetic defense, CSPD or ISPD, that enhance the robustness of detectors against phonetic perturbation.
% Our experimental result demonstrated the effectiveness of both our proposed methods and they reflect the real malicious users' intent.
\end{abstract}

\section{Introduction}

% Malicious users often conduct adversarial attacks on offensive text filtering systems by subtly modifying their toxic texts using some strategies that preserve human readability but confuse the systems.
% In response, researchers have investigated attack strategies and corresponding defenses to mitigate such threats.

As offensive text detection systems have been advanced, malicious users have adopted more sophisticated filtering evasion strategies.
In particular, they have been trying to replace characters or words in hateful texts with alternatives that are pronounced similarly.
Despite requiring substantial linguistic awareness of target languages, malicious users frequently adopt phonetic substitution, which proves to be an effective method of evading detection \cite{badcharacter, cryptext}.
Therefore, researchers have formalized this strategy and proposed defense methods against it \cite{Homoglyphs, perturbation_in_wild}.

However, we identify two key challenges regarding the target language and the proposed defense strategies.
First, existing studies on phonetic substitution attacks have rarely considered Korean.
Because users of phonographic writing systems can often infer the original word from its phonetically perturbed form, such attacks may be more effective in languages like Korean \cite{orthographic_feature}.
Nevertheless, prior research has largely overlooked phonetic perturbations in Korean and instead focused on language-agnostic strategies, such as inserting meaningless words \cite{koreanfool}.

Second, most of the currently proposed defense methods primarily focus on constructing perturbed datasets, while less focused on augmenting additional feature representations.
Specifically, proposed defense methods often rely on fine-tuning methods using datasets specialized for each attack strategy \cite{visual-abuse}. However, those approaches not only incur additional annotation costs but also raise concerns about overfitting to particular attack patterns.

% To tackle these challenges, we propose (1) a PHonetic-Informed Substitution for Hangul, PHISH, and (2) cumulative or integrative semantic-phonetic defense methods, CSPD and ISPD.
To tackle these challenges, we propose (1) a PHonetic-Informed Substitution for Hangul, PHISH, and (2) sequntial or direct Mixed Encoding of Semantic-pHonetic features (seq-MESH, dir-MESH).
PHISH substitutes one or two Korean unit letters per syllable with phonetically similar counterparts using the International Phonetic Alphabet (IPA) and the Korean standard pronunciation rules. Unlike prior strategies, PHISH does not use any characters or special symbols from other languages; instead, it leverages only the Korean character set.
seq-MESH and dir-MESH aim to enhance the robustness of detectors against phonetic perturbation by augmenting phonetic information. Specifically, our methods adopt cross-attention mechanism to incorporate semantic and phonetic information.

To examine the effectiveness of both our proposed attack and defense methods, we conducted experiments on two Korean hate speech datasets: K-HATERS \cite{khaters} and KoLD \cite{kold}.
Specifically, we quantified performance degradation of baseline detectors under our phonetic substitution attack.
%how much the detection performance of baseline detectors degrades on test sets perturbed by our phonetic adversarial substitution attack.
Also, we evaluated detectors equipped with our defense methods on both the original and perturbed test sets, comparing their performance to the corresponding base models.
Thus, this paper has following contributions:

\begin{itemize}
\item We introduce a phonetic substitution attack method, \phish, which leverages the characteristics of the Korean language and successfully misleads prior detectors.
\item Also, we propose defense methods, sequential or direct \mesh, which enhance the robustness of detectors by guiding them to incorporate semantic and phonetic information.
% \item Also, we propose defense methods, CSPD and ISPD, which enhance the robustness of detectors by guiding them to incorporate semantic and phonetic information.
% \item The experimental results demonstrate the effectiveness of our proposed methods: the attack consistently degraded the performance of baseline detectors, while the defense methods improved detection performance on both perturbed and unperturbed test sets.

% \item Also, we propose a defense method, CSPD, which guides transformer-based detectors to incorporate semantic and phonetic information by accumulating self-attention and cross-attention layers.
% \item Additionally, we present a lighter defense method, ISPD, which steers detectors to consider semantic and phonetic information simultaneously using a single cross-attention layer.

% \item Also, we present two defense methods, PAD and C-PAD, which guide transformer-based detectors to incorporate phonetic information by adding or replacing their self-attention layers with cross-attention layers.
\end{itemize}

\section{Attack method}
\label{sec:attack method}
Korean malicious users often circumvent filtering systems by making slight modifications to their toxic sentences.
Specifically, they commonly conduct \textit{phonetic substitution}: replacing offensive letters or words with phonetically similar alternatives.
As Korean is a phonographic language with shallow orthographic depth, phonetically substituted toxic texts remain intelligible to human readers but can easily confuse detection systems that rely on semantic representations \cite{orthographic-depth}.
In the Korean writing system, Hangul, each character fundamentally represents a single syllable. Here, a Hangul syllable character is structured by combining individual components, called \textit{jamo}, into a syllable block. Such a syllable block must contain at least one initial consonant (onset) and a vowel (nucleus), while a final consonant (coda) may or may not be present. %\footnote{In this study, we use the term \textit{syllable} to refer to a single Hangul character, which is formed by combining multiple jamos and corresponds to one syllabic block.}.
For example, a Hangul syllable block `김 [\textipa{kim}]' consists of three jamos, onset `ㄱ [\textipa{k}]', nucleus `ㅣ [\textipa{i}]', and coda `ㅁ [\textipa{m}].'

% Malicious users often circumvent filters by making slight modifications to their toxic sentences.
% Commonly, such users replace letters in a hateful text with others that are visually or phonetically similar, which are called visual substitution and phonetic substitution, respectively.
% For example, the English word `smile' can be substituted for `smi1e' or `smyle.'

% Visual substitution is comparatively straightforward to implement and , as it simply requires specifying visually similar character mappings. However, the range of such substitutions is inherently limited (e.g., ‘o’ to ‘0’, ‘l’ to ‘1’), rendering them susceptible to detection by simple rule-based filtering.
% In contrast, although phonetic substitution requires a deeper understanding of the phonological systems of the target language, it can more effectively evade rule-based filters by introducing more variation in textual form than visual substitution.
% Especially, phonetic substitution is effective in languages with shallow orthographic depth, where characters closely mirror their phonetic realization.
% Therefore, we propose a phoneme-wise adversarial attack method, PAA, that uses the phonetic substitution as a key element.
% Specifically, we mainly target the Korean language because its writing system, Hangul, is a syllabic system with relatively shallow orthographic depth.

\begin{algorithm}[t]
\caption{PHISH Algorithm}\label{alg:attack}
\begin{algorithmic}[1]
\Require Text $\mathcal{T}=\{\mathcal{T}_0,\cdots,\mathcal{T}_n\}$,
\Statex\phantom{Inp} Perturbation ratio $r \in [0, 1]$,
\Statex\phantom{Inp} Attack mode $m \in \{\text{Single},\text{ Dual}\}$
\Ensure Perturbed text $\mathcal{T}$
\Statex
\State $\mathcal{I_D}, \mathcal{I_S} \gets \texttt{Vulnerable Search}(\mathcal{T})$ \label{pasa:vs}
\State Shuffle $\mathcal{I_D}$ and $\mathcal{I_S}$ \label{pasa:shuffle}
\State $n_V \gets $ Total length of $\mathcal{I_D}$ and $\mathcal{I_S}$ \label{pasa:nv}
\State $n_A \gets 0$ \Comment \# of perturbed syllables \label{pasa:na}
\Statex
% \text{attackable indices list } \mathcal{I} \gets \newline

% \State $\text{shuffle}(\mathcal{I_D})$
% \State $\text{shuffle}(\mathcal{I_S})$

\While{$\frac{n_A}{n_V}< r \text{ and }\mathcal{I_D} \neq \varnothing$} \label{pasa:id-loop-start}
\State Pop a target index $i$ from $\mathcal{I_D}$
% $i \gets \text{POP}(\mathcal{I_D})$ \Comment The target index
% \State $sybl \gets \mathcal{T}_i$
\State $\mathcal{T}_i \gets \texttt{Syllable Attack}(\mathcal{T}_i, n_{attk})$ 
% \Statex\Comment Attack a syllable in $\mathcal{I_D}$
\State $n_A \gets n_A+1$
\EndWhile \label{pasa:id-loop-end}

\Statex
\While{$\frac{n_A}{n_V}< r \text{ and }\mathcal{I_S} \neq \varnothing$} \label{pasa:is-loop-start}
\State Pop a target index $i$ from $\mathcal{I_S}$
% \State $i \gets \text{POP}(\mathcal{I_S})$
% \State $sybl \gets \mathcal{T}_i$
\State $\mathcal{T}_i \gets \texttt{Syllable Attack}(\mathcal{T}_i, n_{attk})$
% \Statex\Comment Attack a syllable in $\mathcal{I_S}$
\State $n_A \gets n_A+1$

\EndWhile\label{pasa:is-loop-end}
\State \Return $\mathcal{T}$

% \textbf{Output:}
\end{algorithmic}
\end{algorithm}
Based on this structural property, we propose a PHonetic-Informed Substitution for Hangul (PHISH) that perturbs Korean text to mislead detection systems.
PHISH replaces a subset of jamos within each syllable with phonetically similar alternatives, using the International Phonetic Alphabet (IPA) and the Korean standard pronunciation rule.
In particular, PHISH uses two degrees of attack according to the number of substituted jamos within a syllable: single-jamo attack, where only one jamo is substituted, and dual-jamo attack, where two jamos are substituted. During the attack, PHISH employs a look-up table $\mathcal{D}$ to match phonetically similar jamos.
Section \ref{subsection:pasa} details PHISH algorithm and Section \ref{subsection:dictionary} illustrates how we defined the predefined look-up table $\mathcal{D}$.

% Also, as PAA requires predefined rules to conduct substitutions, we constructed a look-up table $\mathcal{D}$, which is used as replacement rules, that provides a set of phonetically closer jamos to a given jamo. The following sections detail the algorithm of our attack method and the procedure for defining the look-up table $\mathcal{D}$.

% \subsection{Phonetic Adversarial Substitution Attack, PASA}
\subsection{The PHISH algorithm}
\label{subsection:pasa}
\begin{algorithm}[t]

\caption{Vulnerable Search Algorithm}\label{alg:vulnerable-search}
\begin{algorithmic}[1]
\Require Look-up table $\mathcal{D}$, Text $\mathcal{T}$
\Ensure Double-indices list $\mathcal{I_D}$,
\Statex\phantom{Outp} Single-indices list $ \mathcal{I_S}$
\Statex
\For {$\text{each syllable } sybl \text{ in } \mathcal{T}$} 
    \State $c \gets 0$ \Comment \# of substitutable jamos \label{vs:syllable-search-start}
    \For {$\text{each jamo } j \text{ in } sybl$}
    % \State $sybl \gets \mathcal{T}_i$
    % \State $\mathcal{J}=\{\mathcal{J}_0,...,\mathcal{J}_n\} \gets \text{decompose}(\mathcal{T}_i)$ \Comment Decompose a syllable into a list of jamos
    
    % \For{$j \gets 0\text{ to } len(\mathcal{J})$}
        \If{$\mathcal{D}[j] \neq \varnothing$} \Comment Alternatives exist
            \State $c \gets c+1$
        \EndIf
    \EndFor \label{vs:syllable-search-end}
    \Statex
    % \EndFor
    \If{$c\geq2$} \label{vs:append-start}
    % \Statex\Comment More than one substitutable jamos
    \State Add the index of $sybl$ into $\mathcal{I_D}$
    \ElsIf{$c=1$}
    % \Statex\Comment Only one substitutable jamo
    \State Add the index of $sybl$ into $\mathcal{I_S}$
    % \State $\text{append}(\mathcal{I_S},\text{index}(sybl))$
    \EndIf \label{vs:append-end}
\EndFor
\State \Return $\mathcal{I_D}$ and $\mathcal{I_S}$

\end{algorithmic}
\end{algorithm}
Algorithm \ref{alg:attack} shows the pseudocode of the PHISH.
The algorithm takes three inputs: an input text $\mathcal{T}$, which is a sequence of syllables $\mathcal{T}_i$, a perturbation ratio $r$, and the degree of attack $m$. Here, the degrees $m$ of `single' and `dual' refer to single and dual-jamo attacks, respetively.
%Here, the attack mode indicates the number of jamos that will be altered in a syllable. 
% Specifically, when the mode is set as `single', only a jamo will be replaced with another jamo per attacked syllable, while two jamos per syllable for the `dual' mode.

PHISH consists of two main phases: (1) Index searching and (2) Substitution.
In index searching, PHISH identifies the target indices to be perturbed (Line \ref{pasa:vs}) before conducting substitution.
Since some Korean syllables do not allow any perturbation because their jamos do not have any phonetically similar alternatives, the algorithm first identifies the vulnerable indices of $\mathcal{T}$ that allow our adversarial attack. Specifically, if a syllable contains more than one replaceable jamo, its index is added to $\mathcal{I_D}$; otherwise, if it contains only one, the index is added to $\mathcal{I_S}$.
To search this index, PHISH calls `vulnerable search' illustrated in Algorithm \ref{alg:vulnerable-search} (Section \ref{subsection:vs}).

After determining the target indices, the substitution phase starts (Lines \ref{pasa:id-loop-start} to \ref{pasa:is-loop-end}).
In this phase, the algorithm perturbs syllables corresponding to target indices one by one until the ratio of attacked syllables reaches the given ratio $r$ or no more vulnerable indices are left.
For the substitution, PHISH uses syllable attack algorithm, which is illustrated in Section \ref{subsubsection:sa}.
% Note that PASA prioritizes syllables corresponding to $\mathcal{I_D}$ over those in $\mathcal{I_S}$.
% This is because such syllables are more likely to contain a larger number of jamos, meaning that substituting a fixed number of jamos affects a smaller proportion of the syllable. It means that perturbed syllables from $\mathcal{I_D}$ are more likely to be intelligible to human readers by preserving higher perceptual similarity.
After the substitution phase is done, PHISH returns the perturbed text $\mathcal{T}$.

% Note that the PASA aims to syllables corresponding to $\mathcal{I_D}$ prior to those corresponding to the $\mathcal{I_S}$. The reason for this priority is
% to lower the impact of the attack.
% Syllables that allow a dual-jamo attack are more likely to consist of a larger number of jamos compared to syllables that allow only a single-jamo attack. In other words, syllables structured as `cvc', which contain three jamos, are more likely to permit dual-jamo attacks than `cv' syllables containing only two jamos. Therefore, syllables with more jamos experience a relatively lower proportion of replaced jamos when attacked, and the impact of the attack on its overall integrity or readability tends to be smaller.
% When the replacing phase is done, the algorithm returns the attacked version of $\mathcal{T}$.

\subsubsection{Vulnerable Search Algorithm}
\label{subsection:vs}

Algorithm \ref{alg:vulnerable-search} shows the search algorithm for identifying target indices of a given text $\mathcal{T}$.
When $\mathcal{T}$ is inputted, the algorithm checks whether each syllable $sybl$ in $\mathcal{T}$ allows perturbation.
In detail, the algorithm iterates over each syllable in $\mathcal{T}$, and checks whether each jamo composing each syllable has alternatives by referring to a predefined look-up table (Lines \ref{vs:syllable-search-start} to \ref{vs:syllable-search-end}).
When a syllable has substitutable jamo, the index of syllable is appended to $\mathcal{I_D}$ or $\mathcal{I_S}$ according to the number of substitutable jamos (Lines \ref{vs:append-start} to \ref{vs:append-end}).
After the iteration, the algorithm returns the two indices list, $\mathcal{I_D}$ and $\mathcal{I_S}$.

% In detail, it decomposes each syllable into a list of jamos and counts the number of substitutable jamos (Lines 9 to 12). Here, the algorithm uses a predefined table $\mathcal{D}$ to figure out the existence of alternative jamos. %The detail of the table is illustrated in Section \ref{subsection:dictionary}.
% Each index of syllables is classified according to the number of its substitutable jamos (Lines 14 to 17).

% Indices of syllables with multiple substitutable jamos are appended to $\mathcal{I_D}$, while those with exactly one substitutable jamo are appended to $\mathcal{I_S}$.
% After checking all syllables in the text, the algorithm returns $\mathcal{I_D}$ and $\mathcal{I_S}$.

% In the vulnerable index search step, when a text is inputted, the algorithm finds indices of vulnerable syllables that contain more than one vulnerable letter, which can be replaced (Line 3).
% Here, it uses the vulnerable search (VS) algorithm shown in Algorithm \ref{alg:vulnerable-search}.
% The VS algorithm returns two indices list, $\mzathcal{I_D}$ and $\mathcal{I_S}$.
% Syllables located at the indices present in the $\mathcal{I_D}$ within the text contains more than two vulnerable letters, while syllables according to $\mathcal{I_S}$ have only one.

% After the vulnerable searching step, the replacement step starts.
% In this step, the algorithm uses the syllable attack (SA) algorithm to attack syllables one by one until the portion of attacked syllables meets the $r$ or no more unattacked vulnerable syllables are left.
% Lines 6 to 15 of Algorithm \ref{alg:attack} show the replacement step.

\subsubsection{Syllable Attack Algorithm}
\label{subsubsection:sa}
\begin{algorithm}[t]

\caption{Syllable Attack Algorithm}\label{alg:attack-syllable}
\begin{algorithmic}[1]
\Require Look-up table $\mathcal{D}$, Syllable $sybl$,
\Statex\phantom{Inp} Degree of attack $m \in \{\text{Single, Dual}\}$
\Ensure Perturbed syllable $sybl$
\Statex
\State Initialize \# of substitutable jamos $n_{sttd}$ as 0
% \If{$m=\text{Single}$} 
% \State $n_{attk} \gets 1$
% \ElsIf{$m=\text{Dual}$}
% \State $n_{attk} \gets 2$
% \EndIf \label
\State \textbf{if} $m = \text{Single}$ \textbf{then} $n_{attk} \gets 1$ \label{sa:check-degree-start}
\State \textbf{else if} $m = \text{Dual}$ \textbf{then} $n_{attk} \gets 2$ \label{sa:check-degree-end}
\State \textbf{end if}

% \State $\mathcal{J} = \{\mathcal{J}_0,...,\mathcal{J}_n\} \gets \text{decompose}(sybl)$ \Comment Decompose a syllable into a list of jamos
% \State $\mathcal{J} = \{j_0,...,j_n\} \gets \text{decompose}(sybl)$ \label{sa:decompose-jamo}
\State Decompose $sybl$ into a list of jamos $\mathcal{J}$. \label{sa:decompose-jamo}
\State Shuffle list $\mathcal{J}$.
\Statex
% \State $\mathcal{J} \gets \text{shuffle}(\mathcal{J})$
% \State $\text{Shuffle }\mathcal{J}$
% \State $n_{stt} \gets 0$
% \State $\text{shuffle}(\mathcal{J})$

% \For{$i \gets 0 \text{ to } len(\mathcal{J})$}
\For {$\text{ each jamo } j \text{ in } \mathcal{J}$} \label{sa:substitution-start}
    % \If{$\mathcal{D}[\mathcal{J}_i] \neq \varnothing$}
    \If{$\mathcal{D}[j] \neq \varnothing$}
        % \State $\mathcal{J}_i \gets \text{Random Choice}(\mathcal{D}[j])$
        \State Substitute $j$ with random jamo in $\mathcal{D}[j]$
        \State $n_{sttd} \gets n_{sttd} +1$
    \EndIf

    \If{$n_{sttd} = n_{attk}$}
        \State $\textbf{break}$
    \EndIf \label{sa:substitution-end}
\EndFor

% \While{$n_{rpl} < n_{attk}$ and $\mathcal{J} \neq \varnothing$}
%     \State $j \gets \text{POP random}(\mathcal{J})$
%     % \State $C \gets \mathcal{D}[j]$ \Comment Similar letters of letter $j$
%     % \If{$C \neq \varnothing$}
%     \If{$\mathcal{D}[j] \neq \varnothing$} \Comment Vulnerable letter
%         \State $c \gets \text{Random Choice}(C)$ 
%         \State $\text{append}(\mathcal{J}_{attked}, c)$
%         \State $n_{rpl} \gets n_{rpl} +1$
%     \Else
%         \State $\text{append}(\mathcal{J}_{attked}, j)$
%     \EndIf
% \EndWhile
% \State $sybl \gets \text{compose}(\mathcal{J})$
\State Recompose $sybl$ with substituted jamos $\mathcal{J}$
\State \Return $sybl$
\end{algorithmic}
\end{algorithm}
Algorithm \ref{alg:attack-syllable} illustrates the process of attack syllables.
The algorithm requires a syllable to be attacked and the degree of attack.
% If one use the \textit{single} degree of attack, the number of jamos to be substituted $n_{attk}$ is set as 1, while 2 for the \textit{dual} mode (Lines 4 to 7).
After deciding the number of jamos to be substituted (Lines \ref{sa:check-degree-start} to \ref{sa:check-degree-end}), we decompose the inputted syllable $sybl$ into a list of jamos (Line \ref{sa:decompose-jamo}).
% Specifically, the length of the list is two (an initial consonant and a vowel) or three (an initial consonant, a vowel, and a final consonant).
Then, the decomposed list is shuffled to substitute jamos with a random order.
After, the algorithm substitutes each jamo with its phonetically similar alternatives by using the look-up table $\mathcal{D}$ until the number of substituted jamos $n_{stt}$ reaches the predefined threshold $n_{attk}$ (Lines \ref{sa:substitution-start} to \ref{sa:substitution-end}).
After the substitution, the algorithm returns the perturbed syllable, composed of substituted jamos.

\subsection{Look-up Table for Alternatives}
\label{subsection:dictionary}

\begin{figure*}[t]
  \includegraphics[width=\textwidth]{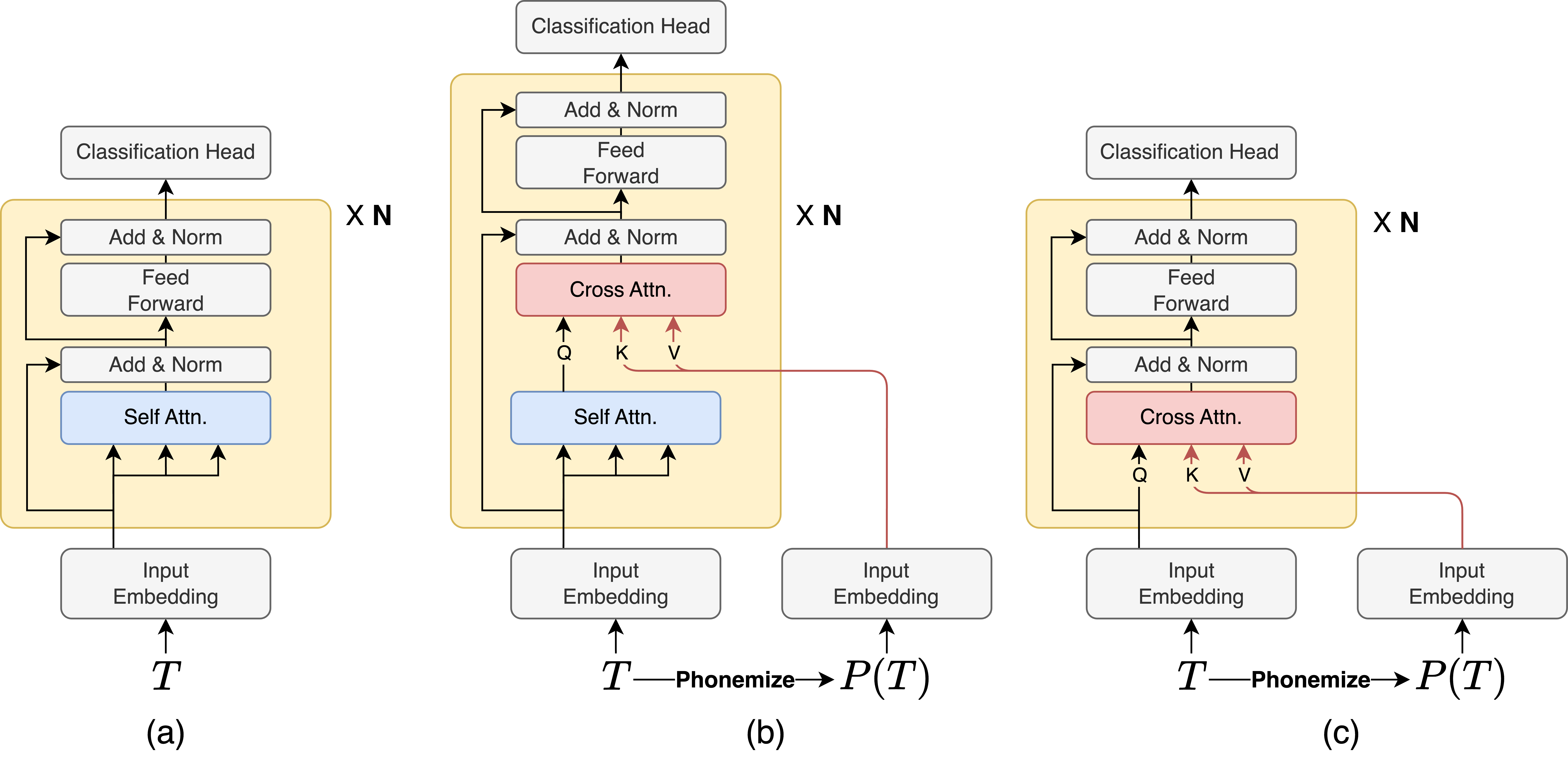}
  \caption{Architectures of base models and our methods. (a) shows the architecture of base detectors using the self-attention mechanism; (b) shows the architecture of seq-MESH detectors using stacked self and cross-attention layers; (c) shows the architecture of dir-MESH detectors using cross-attention instead of self-attention.}
  \label{fig:framework}
\end{figure*}

Our proposed adversarial attack, PHISH, requires a predefined look-up table $\mathcal{D}$ that maps a jamo to a set of phonetically similar jamos.
% Defining this dictionary is crucial for the attack because 
As previously mentioned, a Korean syllable consists of an initial consonant, a medial vowel, and an optional final consonant. Thus, we applied different procedures for each component of syllables when constructing the look-up table. Appendix \ref{app:lookup} illustrates the table. 

% As previously noted, a Korean syllable consists of an initial consonant, a medial vowel, and an optional final consonant. Since consonants vary by position and may exhibit different phonetic realizations depending on their role, we classify jamos by syllabic position and construct sets of phonetically similar jamos based on phonological similarity. 여기!
% Here, we used the International Phonetic Alphabet (IPA) and the Korean standard pronunciation rule as the principles of defining phonetically similar jamos.
% The following paragraphs describe how we defined sets of phonetically similar jamos for each syllabic position—initial consonants, final consonants, and medial vowels.
% See Appendix \ref{app:dictionary} for the detail of the look-up table.
% We classified initial consonants according to their base IPA symbols. As shown in Table \ref{tab:consonant_ipa}, there are some groups of consonants located in the same position of articulation and manner. Also, consonants sharing such positions also use the same base IPA symbols.
% For example, the base symbols of consonants `ㅂ [\textipa{p}]', `ㅃ [\textipa{p*}]', and `ㅍ [\textipa{p\super h}]' are identical to `[\textipa{p}]'.
% While the difference among them is defined by the laryngeal setting, this feature contributes relatively little to their phonetic distinctiveness compared to articulatory properties such as place and manner of articulation.

To classify initial consonants, we used their base IPA symbols as the guiding principle.
% We classified initial consonants based on their base IPA symbols.
% As shown in Table \ref{tab:consonant_ipa}, some consonants share the same place and manner of articulation the same base IPA symbol.
In Korean, some initial consonants share the same place and manner of articulation. We grouped initial consonants sharing similar articulatory features or the base phone.
% Some Korean consonants share the same place and manner of articulation
For example, `ㅂ [\textipa{p}]' and `ㅍ [\textipa{p\super h}]' are variants of the base phone [\textipa{p}]'.
While their differences arise from laryngeal settings, such distinctions contribute less to phonetic similarity than their articulation place and manner.
Accordingly, we defined five sets for the initial consonants regarding their base phone.
% : `p', `t', `k', `\textipa{t\textctc}', and `s.'

When defining the table for final consonants, we used the Korean standard pronunciation rule as the principle. Unlike initial consonants, which are pronounced distinctly from others, some final consonants are pronounced identically according to the Korean standard pronunciation rule.
As this phonological property is aligned with the motivation of PHISH, we directly used this rule to define the table for final consonants.
% Additionally, special complex consonants, which are restricted to be used as final consonants (e.g., `ㄳ' is a combination of two consonants 'ㄱ [k]' and `ㅅ [s]' that only appears as a final consonant), were also classified according to their realized pronunciation.
Following this procedure, we defined six jamo sets corresponding to one of the six standard pronunciations.

To classify the medial vowels, we grouped monophthongs and diphthongs that share the same base phone.
Specifically, there are 11 Korean diphthongs that are derived from monophthongs by combining them with glides such as `/\textipa{w}/' or `/\textipa{j}/.'
Since diphthongs and their corresponding base vowels are pronounced in a similar way, we grouped them into the same substitution sets.
Accordingly, we defined seven sets of vowels.

\section{The MESH Defense Methods}
\label{sec:defense}
We hypothesized that augmenting the phonetic information can enhance the robustness of detectors regarding two aspects against phonetic substitution attacks: (1) providing supplementary features, and (2) mitigating information loss.
First, the augmented phonetic information allows prior detectors, which rely on semantic-level representations, to leverage alternative linguistic cues during their detection.
% First, the augmented phonetic information serves as a supplementary feature, allowing detectors, which primarily focus on semantic-level representations, also to leverage alternative linguistic cues during the detection process.
Since semantics of perturbed texts can significantly deviate from that of original text, such phonetic cues can help to address such deviation.
% help prevent detectors from being severely misled.

Second, augmenting phonetic features can mitigate the information loss caused by perturbations.
As perturbation increases the likelihood of unknown tokens during the tokenization process of detectors, it can severely disrupt the semantic structure of the text \cite{koreanfool}.
This disruption can interfere with the detectors' semantic understanding of the text.
As phonetic information can provide a hint for reconstructing the unknown word, we believe that augmenting phonetic information can recover such information loss.
% However, incorporating features can recover such information loss by augmenting phonetic information for unknown tokens.
% and reduce the misinterpretation by detectors.

To combine phonetic information to detectors, we propose sequential or direct Mixed Encoding of Semantic-pHonetic features (in short, seq-MESH or dir-MESH) that use additional or alternative cross-attention layers. Figure \ref{fig:framework} compares prior detectors and our proposed methods. Previously, detectors use self-attention layers to process or capture the semantic meaning of the input text, as shown in Figure \ref{fig:framework} (a). Meanwhile, seq-MESH and dir-MESH combine the input text and its phoneme sequence using cross-attention layers, as shown in Figure \ref{fig:framework} (b) and (c).
Here, to generate phoneme sequences, we used a widely adopted open-source Korean phonemizer\footnote{\url{https://github.com/Kyubyong/g2pK}}.
\subsection{Sequential MESH}
\label{subsection:pad}
While seq-MESH follows the same overall architecture of previous detectors, it differs by incorporating an additional cross-attention layer in every encoder block.
This additional layer computes attention between the semantics of the input text and its phoneme sequence. Specifically, we used the output of the preceding self-attention output as query; and the embedded phoneme sequence is used for key and value.
As a result, seq-MESH can fully leverage and incorporate the two different types of features using two attention layers.

% \isitnecessary
% CSPD uses two cumulative attention layers for every encoder layer.
% Specifically, CSPD requires an additional cross-attention layer to the right after the self-attention layer for every encoder layer.

% When a text $T$ is input to CSPD, a phoneme sequence $P(T)$ is generated using an external phonemizer\footnote{We used a widely used open-source Korean phonimizer library from \texttt{https://github.com/Kyubyong/g2pK}}.
% The embedding of $P(T)$ is residually connected to every encoder layer to let detectors conduct cross-attention right after each self-attention layer.
% Here, the residually connected embedded $P(T)$ is used as key and value for the cross-attention, while the embedded representation of $T$ or the outputs of preceding layers are used as queries.
% By doing so, detectors can leverage the semantics and phonetics of given texts simultaneously by conducting two attention mechanisms that combine the two different information.
% \neednothing

\subsection{Direct MESH}
\label{subsection:cpad}
Since self-attention layers specialize in capturing and processing the semantics of the text, they may propagate the distorted semantics caused by unknown tokens in perturbed texts.
Specifically, this error propagation of self-attention layers can affect the subsequent layers and mislead the detectors.
To address this, we further propose dir-MESH, which replaces the self-attention layers with cross-attention layers to incorporate semantic and phonetic information directly, while relaxing the possibility of error propagation.
Specifically, we use the same architecture with seq-MESH except for self-attention layers.

\section{Experiment}
\label{sec:experiment}
% In this section, we detail the experiment that we conducted to investigate the effectiveness of both attack and defense methods.
% To assess our attack method, we measured the performance degradation of baseline detectors when we applied PASA on Korean hate speech datasets with different attack settings.
% And, to evaluate our defense methods, we compared the detection performance between baseline detectors and those with CSPD/ISPD methods.

% the detection performance of detectors adopting CSPD or ISPD with their corresponding base models under perturbed datasets.
% Specifically, we attacked the test sets of two Korean hate speech datasets under various conditions and measured the performance drop of the detectors.
% To assess the effectiveness of our defense methods, we compared the detection performance of base models and their augmented counterparts--using CSPD or ISPD-- on both the original (non-attacked) and adversarially perturbed test sets, evaluating improvements in both general detection performance and adversarial robustness.
% Also, we assessed our defense methods by comparing the detection performance of baselines, which use the self-attention mechanism, with the performance of the modified models that use PAD or C-PAD instead of self-attention.
% Also, we assessed our defense methods by comparing the detection performance of tested models that adopt three different architectures shown in Figure \ref{fig:framework}.
% The following subsections detail the datasets and implementations of baselines.

\subsection{Datasets}
\label{subsection:datasets}
\begin{table*}[t] %[!htp]
\centering
% \resizebox{\textwidth}{!}{%
% Please add the following required packages to your document preamble:
% \usepackage{graphicx}
% \usepackage[table,xcdraw]{xcolor}
% Beamer presentation requires \usepackage{colortbl} instead of \usepackage[table,xcdraw]{xcolor}
% Please add the following required packages to your document preamble:
% \usepackage{graphicx}
% \usepackage[table,xcdraw]{xcolor}
% Beamer presentation requires \usepackage{colortbl} instead of \usepackage[table,xcdraw]{xcolor}
% Please add the following required packages to your document preamble:
% \usepackage{graphicx}
% \usepackage[table,xcdraw]{xcolor}
% Beamer presentation requires \usepackage{colortbl} instead of \usepackage[table,xcdraw]{xcolor}
% \begin{tabular}{lr|r@{$\pm$}lr@{$\pm$}lr@{$\pm$}lr@{$\pm$}lr@{$\pm$}lr@{$\pm$}l}
\begin{tabular}{lr@{$\pm$}l|r@{$\pm$}l@{\;\;\;}r@{$\pm$}l|r@{$\pm$}l@{\;\;\;}r@{$\pm$}l|r@{$\pm$}l@{\;\;\;}r@{$\pm$}l}
\toprule
Attack Ratio                  & \multicolumn{2}{c}{0\%}             & \multicolumn{4}{c}{10\%}            & \multicolumn{4}{c}{20\%}    & \multicolumn{4}{c}{30\%}\\ 
& \multicolumn{2}{c}{F1}             & \multicolumn{2}{c}{F1} & \multicolumn{2}{c}{$\Delta$F1}            & \multicolumn{2}{c}{F1}            & \multicolumn{2}{c}{$\Delta$F1}     &   \multicolumn{2}{c}{F1} & \multicolumn{2}{c}{$\Delta$F1}\\

\midrule
BERT             & 73.8 & 0.2       & 73.6 & 0.3    & -0.2 & 0.4     & 71.9 & 0.3        & -1.9 & 0.4   & 69.5 & 0.3   & -4.3 & 0.4        \\
RoBERTa           & 65.0 & 2.0       & 63.5 & 2.6    & -1.5 & 3.3     & 58.2 & 3.3        & -6.8 & 3.9   & 52.4 & 4.6   & -12.6 & 5.0        \\
KCBERT                & 76.2 & 0.4       & 75.7 & 0.3    & -0.5 & 0.5     & 74.8 & 0.3        & -1.4 & 0.5   & 73.4 & 0.3   & -2.8 & 0.5         \\
% Roberta-base           & 77.2      & 76.8&3.24      & 80.0&2.81     & 74.7&1.77      & 77.1&2.13\\
%SeqXGPT*               & 54.1      & 86.5&0.48      & 45.9&0.23     & 41.6&0.31      & 42.3&0.52                                 \\

\midrule
BERT\dirmesh           & 74.2 & 0.5       & 73.1 & 0.4    & -1.1 & 0.6     & 71.7 & 0.5        & -2.5 & 0.7   & 70.2 & 0.8   & -4.0 & 0.9 \\
RoBERTa\dirmesh            & 74.4 & 0.4       & 72.9 & 0.5    & -1.3 & 0.6     & 71.2 & 0.8        & -3.2 & 0.9   & 69.9 & 0.7   & -4.5 & 0.8          \\
KCBERT\dirmesh     & 76.6 & 0.4       & 75.5 & 0.5    & -1.1 & 0.6     & 74.2 & 0.5        & -2.4 & 0.6   & 73.2 & 0.6   & -3.4 & 0.7          \\

\midrule
BERT\seqmesh            & 78.9 & 0.4       & 76.2 & 0.5    & -2.7 & 0.6     & 73.0 & 0.5        & -5.9 & 0.6   & 70.8 & 0.4   & -8.1 & 0.6 \\
RoBERTa\seqmesh            & 74.6 & 0.6       & 73.4 & 0.6    & -1.2 & 0.8     & 71.4 & 0.8        & -3.2 & 1.0   & 70.1 & 0.9   & -4.5 & 1.1          \\
KCBERT\seqmesh     & \textbf{80.8} & \textbf{0.2}       & \textbf{79.2} & \textbf{0.3}    & -1.6 & 0.4     & \textbf{77.8} & \textbf{0.3}        & -3.0 & 0.3   & \textbf{74.9} & \textbf{0.4}   & -5.9 & 0.4          \\

\bottomrule
% \multicolumn{10}{r}{\small * Models used GPT-2 as a proxy model, except Llama 3.}
\end{tabular}%
% }
\caption{Detection performance on K-HATERS dataset with single-jamo attack}
\label{tab:khaters-single-result-full}
\end{table*}
\begin{table*}[h!] %[!htp]
\centering
% \resizebox{\textwidth}{!}{%
% Please add the following required packages to your document preamble:
% \usepackage{graphicx}
% \usepackage[table,xcdraw]{xcolor}
% Beamer presentation requires \usepackage{colortbl} instead of \usepackage[table,xcdraw]{xcolor}
% Please add the following required packages to your document preamble:
% \usepackage{graphicx}
% \usepackage[table,xcdraw]{xcolor}
% Beamer presentation requires \usepackage{colortbl} instead of \usepackage[table,xcdraw]{xcolor}
% Please add the following required packages to your document preamble:
% \usepackage{graphicx}
% \usepackage[table,xcdraw]{xcolor}
% Beamer presentation requires \usepackage{colortbl} instead of \usepackage[table,xcdraw]{xcolor}
% \begin{tabular}{lr|r@{$\pm$}lr@{$\pm$}lr@{$\pm$}lr@{$\pm$}lr@{$\pm$}lr@{$\pm$}l}
\begin{tabular}{lr@{$\pm$}l|r@{$\pm$}l@{\;\;\;}r@{$\pm$}l|r@{$\pm$}l@{\;\;\;}r@{$\pm$}l|r@{$\pm$}l@{\;\;\;}r@{$\pm$}l}
\toprule
Attack Ratio                  & \multicolumn{2}{c}{0\%}             & \multicolumn{4}{c}{10\%}            & \multicolumn{4}{c}{20\%}    & \multicolumn{4}{c}{30\%}\\ 
& \multicolumn{2}{c}{F1}             & \multicolumn{2}{c}{F1} & \multicolumn{2}{c}{$\Delta$F1}            & \multicolumn{2}{c}{F1}            & \multicolumn{2}{c}{$\Delta$F1}     &   \multicolumn{2}{c}{F1} & \multicolumn{2}{c}{$\Delta$F1}\\ 
\midrule
BERT             & 75.1 & 0.5       & 74.6 & 0.6    & -0.5 & 0.8     & 70.6 & 0.7        & -4.5 & 0.9   & 66.4 & 1.5   & -8.7 & 1.6        \\
RoBERTa          & 72.6 & 1.6       & 71.6 & 1.7    & -1.0 & 2.3     & 66.6 & 3.0        & -6.0 & 3.4   & 60.8 & 5.8   & -11.8 & 6.0        \\
KCBERT           & 77.5 & 0.4       & 76.7 & 0.6    & -0.8 & 0.7     & 75.4 & 0.7        & -2.1 & 0.8   & 72.6 & 1.3   & -4.9 & 1.4        \\
% Roberta-base           & 77.2      & 76.8&3.24      & 80.0&2.81     & 74.7&1.77      & 77.1&2.13\\
%SeqXGPT*               & 54.1      & 86.5&0.48      & 45.9&0.23     & 41.6&0.31      & 42.3&0.52                                 \\

\midrule
BERT\dirmesh            & 75.9 & 0.5       & 75.0 & 0.7    & -0.9 & 0.9     & 73.0 & 0.6        & -2.9 & 0.8   & 71.6 & 0.8   & -4.3 & 0.9 \\
RoBERTa\dirmesh            & 75.9 & 0.7       & 75.1 & 0.6    & -0.8 & 0.9     & 73.6 & 0.4        & -2.3 & 0.8   & 71.3 & 0.6   & -4.6 & 0.9          \\
KCBERT\dirmesh     & 77.7 & 0.5       & 76.8 & 0.6    & -0.9 & 0.8     & 74.9 & 0.6        & -2.8 & 0.8   & 73.9 & 0.6   & -3.8 & 0.8          \\

\midrule
BERT\seqmesh            & 79.5 & 1.0       & 77.9 & 0.9    & -1.6 & 1.3     & 74.8 & 0.9        & -4.7 & 1.3   & 73.0 & 0.9   & -0.65 & 1.3          \\
RoBERTa\seqmesh            & 75.9 & 0.5       & 74.9 & 0.3    & -1.0 & 0.6     & 73.3 & 0.4        & -2.6 & 0.6   & 71.4 & 0.5   & -4.5 & 0.7          \\
KCBERT\seqmesh     & \textbf{81.4} & \textbf{0.5}       & \textbf{80.3} & \textbf{0.6}    & -1.1 & 0.8     & \textbf{78.9} & \textbf{0.7}        & -2.5 & 0.9   & \textbf{76.2} & \textbf{0.7}   & -5.2 & 0.9          \\ 
\bottomrule
% \multicolumn{10}{r}{\small * Models used GPT-2 as a proxy model, except Llama 3.}
\end{tabular}%
% }
\caption{Detection performance on KoLD dataset with single-jamo attack}
\label{tab:kold-single-result-full}
\end{table*}
We used two Korean hate speech datasets for our experiment: K-HATERS \cite{khaters} and KoLD \cite{kold}.
Both datasets used online comments to crawl hate speech and labeled them.
Specifically, K-HATERS used a Korean online news platform as the source. They labeled hate speech into various sub-categories, regarding the intensity of hatefulness.
Similarly, KoLD crawled the same platform and YouTube to construct the dataset. KoLD used labels different from K-HATERS for offensive samples.

As our study aims to examine the effectiveness of phonetic methods on hateful speech, we decided to focus on coarse labels: \textit{offensive} or \textit{normal}.
Though two datasets provided detailed labels, we gathered fine-grained offensive labels into a single category.
Since this gathering process produced highly imbalanced regarding these two labels, we downsampled the datasets.
Consequently, we used 104,112 samples from K-HATERS and 40,429 samples from KoLD. 
These samples are split into training, validation, and test sets with a ratio of 8:1:1.

% , while KoLD additionally used YouTube as the source.
% K-HATERS contains samples that consist of text and label pairs. Specifically, the types of labels of hateful texts are divided into various sub-categories according to the intensity of hatefulness, such as `offensive,' and `L-1 hate.'
% Likewise, KoLD offers detailed categorized labels for offensive samples.
% Because our study aims to distinguish hateful texts, we merged those subcategories of offensive labels into an `offensive' label. 

% As the resulting datasets are highly imbalanced, we downsampled the datasets.
% As a result, we collected 104,112 samples from K-HATERS and 40,429 samples from KoLD, and split each dataset into training, validation, and test sets with a ratio of 8:1:1.
% Table \ref{tab:dataset} shows the statistics of the collected dataset.
After collecting datasets for our experiment, we collected additional perturbed test sets.
Using PHISH, we derived different test sets with different settings, including attack ratios and degrees.
Specifically, we conducted attacks under three perturbation ratios (10, 20, and 30\%) and two degrees of attack (single-jamo and dual-jamo). 
Note that we did not alter training set; all methods are trained on the original data without applying PHISH.

\subsection{Baselines and MESH variants}
For baselines, we used three small language models that are commonly used in prior Korean hate speech detection research and adopt the self-attention mechanism: KLUE-BERT, KLUE-RoBERTa \cite{klue}, and KCBERT \cite{kcbert}.
These three models possess Korean language understanding capabilities. Specifically, KLUE-BERT and RoBERTa was pretrained on KLUE dataset \cite{klue}, which is a Korean language understanding benchmark. Meanwhile, KCBERT was primarily trained on web-based data such as news articles and user comments. So, it tends to exhibit stronger baseline performance in tasks related to hate-speech detection compared to the other two.

% KLUE-BERT and RoBERTa possess Korean language understanding capabilities by being pretrained on Korean corpus datasets.
% KCBERT was also pretrained on Korean corpus datasets like the other two models. Meanwhile, since KCBERT was primarily trained on web-based data such as news articles and user comments, it tends to exhibit stronger baseline performance in tasks related to hate-speech detection compared to the other two models.

For implementing detectors equipped with seq-MESH or dir-MESH, we reused the parameters as in \citet{bertgeneration}. We set initial parameters of two methods by copying that of the base models, rather than initializing from scratch. Specifically, the self-attention weights of the base models were copied into cross-attention weights of two methods.
% In implementing the PAD and C-PAD models, we reused as many parameters from the baseline models as possible. For instance, the weight matrices required for cross-attention —such as $W^Q, W^K, W^V$ for generating query, key, and value vectors—were not initialized from scratch but instead inherited directly from the pretrained baseline models.

% Instead of reinitializing, PAD and C-PAD reuse attention-related parameters, including query/key/value projection weights, from the baseline model.
% Here, the parameters of PAD or C-PAD models were initialized with the pre-trained parameters of their corresponding baseline models if 
% PAD and C-PAD models were initialized with the pre-trained parameters of their corresponding baseline models, rather than being initialized from scratch.

\subsection{Environment of Experiment}
We used a single RTX A6000 for training and evaluating the models. We trained each model for five epochs with a learning rate of $10^{-5}$ and a batch size of 32. Then, we chose checkpoints with the highest F1 score on the validation set. We repeated each dataset experiment 10 times with different random seeds to ensure reproducibility.

\section{Result and Discussion}

In this section, we present our experimental results, which are shown in Tables \ref{tab:khaters-single-result-full}, \ref{tab:kold-single-result-full}, \ref{tab:khaters-double-result-full}, and \ref{tab:kold-double-result-full}. Tables display the average and standard deviation of F1 scores across the ten experiments. We found three findings of our methods: (1) degradation of performance under PHISH, (2) robustness of seq-MESH and dir-MESH against the attack scenario, and (3) the alignment between real-world scenarios and our attack and defense methods.

% the real-world relevance of our attack and defense methods.
% First, we evaluate the performance degradation of three baseline models under PASA in the single-jamo attack setting to validate the effectiveness of the proposed attack.
% Second, we compare the detection performance of our defense methods with their corresponding base models on perturbed test sets across varying attack ratios.
% Third, we examine whether the proposed defense methods can also improve performance on original (unperturbed) test sets, thereby confirming their general effectiveness beyond adversarial scenarios.
\begin{table*}[t] %[!htp]
\centering
% \resizebox{\textwidth}{!}{%
% Please add the following required packages to your document preamble:
% \usepackage{graphicx}
% \usepackage[table,xcdraw]{xcolor}
% Beamer presentation requires \usepackage{colortbl} instead of \usepackage[table,xcdraw]{xcolor}
% Please add the following required packages to your document preamble:
% \usepackage{graphicx}
% \usepackage[table,xcdraw]{xcolor}
% Beamer presentation requires \usepackage{colortbl} instead of \usepackage[table,xcdraw]{xcolor}
% Please add the following required packages to your document preamble:
% \usepackage{graphicx}
% \usepackage[table,xcdraw]{xcolor}
% Beamer presentation requires \usepackage{colortbl} instead of \usepackage[table,xcdraw]{xcolor}
% \begin{tabular}{lr|r@{$\pm$}lr@{$\pm$}lr@{$\pm$}lr@{$\pm$}lr@{$\pm$}lr@{$\pm$}l}
\begin{tabular}{lr@{$\pm$}l|r@{$\pm$}l@{\;\;\;}r@{$\pm$}l|r@{$\pm$}l@{\;\;\;}r@{$\pm$}l|r@{$\pm$}l@{\;\;\;}r@{$\pm$}l}
\toprule
Attack Ratio                  & \multicolumn{2}{c}{0\%}             & \multicolumn{4}{c}{10\%}            & \multicolumn{4}{c}{20\%}    & \multicolumn{4}{c}{30\%}\\ 
& \multicolumn{2}{c}{F1}             & \multicolumn{2}{c}{F1} & \multicolumn{2}{c}{$\Delta$F1}            & \multicolumn{2}{c}{F1}            & \multicolumn{2}{c}{$\Delta$F1}     &   \multicolumn{2}{c}{F1} & \multicolumn{2}{c}{$\Delta$F1}\\ 
\midrule
BERT             & 73.8 & 0.2       & 73.1 & 0.4    & -0.7 & 0.4     & 69.6 & 0.4        & -4.2 & 0.4   & 66.3 & 0.7   & -7.5 & 0.7        \\
RoBERTa           & 65.0 & 2.0       & 58.2 & 4.0    & -6.8 & 4.5     & 44.9 & 6.2        & -20.1 & 6.5   & 31.5 & 7.6   & -33.5 & 7.9        \\
KCBERT                & 76.2 & 0.4       & 75.1 & 0.3    & -1.1 & 0.5     & 72.5 & 0.2        & -3.7 & 0.4   & 70.6 & 0.3   & -5.6 & 0.5         \\
% Roberta-base           & 77.2      & 76.8&3.24      & 80.0&2.81     & 74.7&1.77      & 77.1&2.13\\
%SeqXGPT*               & 54.1      & 86.5&0.48      & 45.9&0.23     & 41.6&0.31      & 42.3&0.52                                 \\

\midrule
BERT\dirmesh            & 74.2 & 0.5       & 72.6 & 0.6    & -1.6 & 0.8     & 69.7 & 0.6        & -4.5 & 0.8   & 67.6 & 1.0   & -6.6 & 1.1 \\
RoBERTa\dirmesh            & 74.4 & 0.4       & 72.2 & 0.7    & -2.2 & 0.8     & 68.9 & 0.8        & -5.5 & 0.9   & 67.3 & 1.0   & -7.1 & 1.1          \\
KCBERT\dirmesh     & 76.6 & 0.4       & 74.9 & 0.3    & -1.7 & 0.5     & 72.4 & 0.5        & -4.2 & 0.6   & 71.1 & 0.6   & -5.5 & 0.7          \\

\midrule
BERT\seqmesh            & 78.9 & 0.4       & 75.5 & 0.4    & -3.4 & 0.6     & 71.9 & 0.4        & -7.0 & 0.6   & 69.6 & 0.8   & -9.3 & 0.9          \\
RoBERTa\seqmesh           & 74.6 & 0.6     & 72.8 & 0.6     & -1.8 & 0.8         & 69.7 & 0.8   & -4.9 & 1.0   & 67.7 & 0.9       & -6.9 & 1.1    \\
KCBERT\seqmesh     & \textbf{80.8} & \textbf{0.2}       & \textbf{77.7} & \textbf{0.3}    & -3.1 & 0.4     & \textbf{73.8} & \textbf{0.4}        & -7.0 & 0.4   & \textbf{71.6} & \textbf{0.7}   & -9.2 & 0.7          \\
\bottomrule
% \multicolumn{10}{r}{\small * Models used GPT-2 as a proxy model, except Llama 3.}
\end{tabular}%
% }
\caption{Detection performance on K-HATERS dataset with dual-jamo attack}
\label{tab:khaters-double-result-full}
\end{table*}
\begin{table*}[h!] %[!htp]
\centering
% \resizebox{\textwidth}{!}{%
% Please add the following required packages to your document preamble:
% \usepackage{graphicx}
% \usepackage[table,xcdraw]{xcolor}
% Beamer presentation requires \usepackage{colortbl} instead of \usepackage[table,xcdraw]{xcolor}
% Please add the following required packages to your document preamble:
% \usepackage{graphicx}
% \usepackage[table,xcdraw]{xcolor}
% Beamer presentation requires \usepackage{colortbl} instead of \usepackage[table,xcdraw]{xcolor}
% Please add the following required packages to your document preamble:
% \usepackage{graphicx}
% \usepackage[table,xcdraw]{xcolor}
% Beamer presentation requires \usepackage{colortbl} instead of \usepackage[table,xcdraw]{xcolor}
% \begin{tabular}{lr|r@{$\pm$}lr@{$\pm$}lr@{$\pm$}lr@{$\pm$}lr@{$\pm$}lr@{$\pm$}l}
\begin{tabular}{lr@{$\pm$}l|r@{$\pm$}l@{\;\;}r@{$\pm$}l|r@{$\pm$}l@{\;\;}r@{$\pm$}l|r@{$\pm$}r@{\;\;}r@{$\pm$}r}
\toprule
Attack Ratio                  & \multicolumn{2}{c}{0\%}             & \multicolumn{4}{c}{10\%}            & \multicolumn{4}{c}{20\%}    & \multicolumn{4}{c}{30\%}\\ 
& \multicolumn{2}{c}{F1}             & \multicolumn{2}{c}{F1} & \multicolumn{2}{c}{$\Delta$F1}            & \multicolumn{2}{c}{F1}            & \multicolumn{2}{c}{$\Delta$F1}     &   \multicolumn{2}{c}{F1} & \multicolumn{2}{c}{$\Delta$F1}\\ 
\midrule
BERT             & 75.1 & 0.5       & 73.5 & 0.7    & -1.6 & 0.9     & 67.2 & 1.7        & -7.9 & 1.8   & 56.6 & 3.5   & -18.5 & 3.5        \\
RoBERTa           & 72.6 & 1.6       & 69.7 & 2.5    & -2.9 & 3.0     & 56.5 & 8.3        & -16.1 & 8.5   & 41.6 & 13.6   & -31.0 &  13.7       \\
KCBERT                & 77.5 & 0.4       & 76.2 & 0.5    & -1.3 & 0.6     & 74.0 & 1.0        & -3.5 & 1.1   & 69.2 & 2.5   & -8.3 & 2.5         \\
% Roberta-base           & 77.2      & 76.8&3.24      & 80.0&2.81     & 74.7&1.77      & 77.1&2.13\\
%SeqXGPT*               & 54.1      & 86.5&0.48      & 45.9&0.23     & 41.6&0.31      & 42.3&0.52                                 \\

\midrule
BERT\dirmesh            & 75.9 & 0.5       & 74.5 & 0.9    & -1.4 & 1.0     & 71.1 & 1.0        & -4.8 & 1.1   & 68.9 & 1.3   & -7.0 & 1.4 \\
RoBERTa\dirmesh            & 75.9 & 0.7       & 74.3 & 0.7    & -1.6 &  1.0    & 70.8 & 0.7        & -5.1 & 1.0   & 69.6 & 0.9   & -6.3 & 1.1          \\
KCBERT\dirmesh     & 77.7 & 0.5       & 76.0 & 0.7    & -1.7 & 0.9     & 74.1 & 0.5        & -3.6 & 0.7   & 72.4 & 0.8   & -5.3 & 0.9          \\

\midrule
BERT\seqmesh            & 79.5 & 1.0       & 77.6 & 1.3    & -1.9 & 1.6     & 73.7 & 1.4        & -5.8 & 1.7   & 70.9 & 2.3   & -8.6 & 2.5          \\
RoBERTa\seqmesh            & 75.9 & 0.5       & 74.6 & 0.4    & -1.3 & 0.6     & 70.6 & 0.8        & -5.3 & 0.9   & 69.3 & 0.8   & -6.6 & 0.9          \\
KCBERT\seqmesh     & \textbf{81.4} & \textbf{0.5}       & \textbf{79.6} & \textbf{0.5}    & -1.8 & 0.7     & \textbf{76.4} & \textbf{0.7}        & -0.5 & 0.9   & \textbf{72.7} & \textbf{0.7}   & -8.7 & 0.9          \\ 
\bottomrule
% \multicolumn{10}{r}{\small * Models used GPT-2 as a proxy model, except Llama 3.}
\end{tabular}%
% }
\caption{Detection performance on KoLD dataset with dual-jamo attack}
\label{tab:kold-double-result-full}
\end{table*}

First, we quantified the performance degradation of three base models under PHISH using different attack settings to validate its effectiveness.
The experimental result shows that the F1 scores of all base models declined approximately as the attack ratio increased, regardless of the dataset.
Specifically, with the 30\% attack ratio using single-jamo attack, KCBERT's F1 scores decreased by 2.8 and 4.9 points on the K-HATERS and KoLD datasets, respectively, while BERT and RoBERTa showed larger drops ranging from 4.3 to 12.6 points.
Moreover, since the dual-jamo attack perturbs more jamos per syllable than the single-jamo attack, it led to greater performance degradation on the perturbed datasets.
For instance, with a 20\% attack ratio, BERT and RoBERTa showed F1 score drops on the K-HATERS dataset of 4.2 and 20.1, respectively. These degradations are significantly larger than the 1.9 and 6.8 decrement observed under the single-jamo attack with the same attack ratio.

% This result demonstrates that PASA successfully misled base models by distorting the semantic meaning of text.
We suspect this effectiveness stems from the semantic distortion that PHISH made.
Specifically, PHISH may increase the likelihood of unknown tokens during the tokenization process in detectors, which can lead to the omission of the semantic content of texts.
Also, in some cases, the perturbed syllables may have been converted into homophones, which could have partially altered the semantic interpretation of the sentence.
Appendix \ref{app:statistics} details the statistics of unknown tokens of tokenized texts of each detector and provides additional discussion.

% We interpret this performance degradation as primarily resulting from the increased rate of unknown (UNK) tokens introduced by our attack during tokenization.

Second, we compared the performance of detectors using seq-MESH or dir-MESH with their corresponding base models on perturbed test sets.
While base models struggled to identify perturbed offensive texts, detectors incorporating seq-MESH or dir-MESH consistently outperformed their base counterparts.
This trend became more pronounced as the perturbation ratio or attack degree increased.
For example, when the KoLD dataset was attacked with a 10\% single-jamo perturbation, the performance gaps between the base BERT (74.6\%) and its dir-MESH and seq-MESH variants (75.0 and 77.9) were 0.4 and 3.3 F1 points, respectively.
Under a stronger 30\% dual-jamo attack, these gaps increased to 13.3 and 14.3 points: 56.6, 68.9, and 70.9\% for those three models.

These results indicate that our defense methods enhance robustness against phonetic perturbations since they use complementary information.
Such complementary information is not only useful in recovering semantic loss but also improving the overall detection performance.
Specifically, KCBERT\seqmesh\ outperformed other models including its base model, though KCBERT had already been pretrained on online comments and exhibited strong baseline performance.
We believe that such further improvement demonstrates complementary benefits of our methods.

Lastly, we tested whether our methods realistically capture perturbations observed in real-world data. By evaluating their performance on original test sets (0\% attack), the result showed that seq-MESH showed higher performance than their corresponding base models. Specifically, on KoLD dataset, KCBERT\seqmesh\ achieved 81.4\% F1 score, which is 3.9\% higher than its base model. 

These improvements indicate that our assumption of phonetic perturbation is present in the real world. We assumed that malicious users adopt phonetic substitutions to deceive detectors. And, the improvement of our defense methods on original test sets supports this; the real-world dataset may contain such phonetic substitutions, as our method improves the detection performance. So, we conclude that our methods seem to align with the strategies of real-world malicious users.

\section{Background}
% In this section, we provide an overview of adversarial attack and defense strategies in offensive language detection.
% Section \ref{subsection:background-adversarial attack} reviews general approaches across various languages, followed by studies that focus specifically on the Korean language.
% Section \ref{subsection:background-defense} introduces dataset- and model-level defense methods, including those that incorporate linguistic features unique to Korean.

\subsection{Textual Perturbation Attack}
\label{subsection:background-adversarial attack}
% As hate-speech detection systems have advanced, malicious users have begun to employ increasingly diverse strategies to circumvent them\cite{}.

% 필터링 우회 방법에 초점?
As malicious users have been attempting to conduct more sophisticated filtering evasion methods, such as visual or phonetic substitutions, researchers have attempted to formalize such strategies \cite{global-visual, Phoneticdetection}.
% As offensive text detection systems have advanced, malicious users have been attempting to bypass detectors by introducing slight modifications, which are easily interpretable by humans but mislead filter systems, to their toxic texts\cite{Homoglyphs}.
% Specifically, they often replace characters or words in texts with visually similar symbols or manipulate phonetic representations.
% In response, researchers have attempted to formalize and experimentally analyze such adversarial attack strategies\cite{Phoneticdetection, global-visual}.
For example, \citet{global-visual} summarized 12 obfuscation strategies based on a user study and analyzed the impact of these strategies across diverse datasets using ten detection models.
\citet{Phoneticdetection} profiled hate speech spreaders using the frequencies of lexical and phonetic features from their texts.

% \citet{Phoneticdetection} employed lexical (e.g., word and hashtag frequencies) and phonetic (e.g., IPA-based syllables and phonetic frequencies) features to classify hate speech spreaders on Twitter.
% \cite{Phoneticdetection} investigated the detection of phonetically similar variants of profanities in English and Spanish.

Since such adversarial attacks are not universally applicable across all languages due to differences in features such as writing systems, it is crucial to account for language-specific constraints.
For example, visual substitution strategies are not applicable to the Korean language because Unicode encoding does not support the replacement of Hangul jamo with visually-similar non-Hangul characters.
% since Unicode does not support the combination of Hangul jamo with non-Hangul characters into a valid Korean character.
So, researchers have investigated more language-specific adversarial attacks designed explicitly for the Korean language system \cite{koas, transposedletter, koreanfool}.
For example, to reflect the diverse forms of offensive language used by real-world users, \citet{koas} augments training data by using multiple tokenizers. % (Mecab and Komoran).
% \cite{koreangoogle} organized intentional text variations based on various criteria, such as phonetic similarities and transposed-letter similarity effects \cite{transposedletter}.
\citet{koreanfool} proposed adversarial attack strategies, such as inserting, copying, and decomposing, that are commonly adopted by Korean malicious users.
However, these studies did not explore phonetic substitution despite its effectiveness and applicability, as we verified in our experiment.

\subsection{Defense Against Textual Perturbations}
\label{subsection:background-defense}
To defend against textual perturbations conducted by malicious users, researchers have proposed strategy-specific datasets \cite{Homoglyphs, visual-abuse, visualperturbations, portuguese} or model architectural methods.
Regarding datasets, \citet{portuguese} curated a profanity-annotated dataset from Portuguese online comments, identifying 17 obfuscation strategies including phonetic and symbolic substitutions. Also, \citet{visual-abuse} constructed a phishing email dataset incorporating visual perturbations and demonstrated a detection method using CharacterBERT \cite{character-bert}.
However, these methods require manually crafted datasets to train defense methods. Also, fine-tuning on a specific perturbation may cause overfitting on the perturbation.
Meanwhile, our defense method took different approach from these studies. Specifically, our method do not require any additional datasets for phonetic perturbations; rather, we showed that training on a real-world training set without any phonetic attack is enough to achieve good detection performance.

% However, constructing datasets requires human annotators and additional labor costs. Moreover, fine-tuning detectors on datasets that are specifically crafted based on specific attack strategies may lead to overfitting.
% Meanwhile, our defense methods do not require any additional human annotation and have demonstrated their effectiveness on the original test set from the real world.

Some researchers have aimed to propose defense methods in perspective of detector architecture \cite{tocp, koreanfool, bagofphonetic, korean-embeddinglstm}.
For example, \citet{koreanfool} leveraged layer pooling methods to enhance the robustness of detectors against textual perturbations.
\citet{korean-embeddinglstm} proposed an embedding model to address misbehaviors of detectors caused by morphologically similar words.
Since these approaches rely solely on input text, they may lack robustness against phonetic perturbations that cause semantic distortion.
In contrast, our defense address semantic distortion by supplementing the input with phonetic features. Enabling detectors to integrate them as additional information, our method demonstrated strong performance gain.

\section{Conclusion}
In this paper, we suggested \phish, a phonetic substitution attack method tailored for the Korean language. Also, we proposed \mesh, two defense mechanisms designed to enhance robustness against such phonetic perturbations. PHISH exploits the structural and phonographic characteristics of Hangul; the attack method substitutes one or two jamos per syllable with phonetically similar alternatives, using a predefined IPA-based look-up table. Meanwhile, our defense methods incorporate phoneme-level features through cross-attention mechanisms to integrate semantic representations with phonetic information.

Experimental results on two Korean hate speech datasets demonstrated the effectiveness of PHISH in degrading the performance of baseline detectors, validating its adversarial potential. Furthermore, detectors equipped with seq-MESH or dir-MESH consistently outperformed their base models across both perturbed and original test sets, suggesting that our defense methods not only improve robustness but also can be generalized to real-world data where phonetic substitutions may naturally occur.

These findings suggest that phonetic perturbation is a practically relevant and realistic threat in Korean text processing, and that integrating phonetic information into model architectures can mitigate semantic distortion and thus improve detection performance. We hope our work encourages further exploration of language-specific perturbation strategies and architectural defenses that go beyond dataset-level solutions.
\section{Limitations}
Despite the effectiveness of our methods, this paper has three limitations.
%
% First, 제안한 음소 기반 적대적 공격 기법(PAA)은 모든 언어에 보편적으로 적용되기 어렵다. 특히, 본 연구는 음소 단위의 치환이 가능한 한글의 음절 기반 체계와 얕은 음운 깊이에 기반하고 있으며, 이러한 언어적 특성이 없는 일본어, 중국어와 같은 언어에서는 본 공격 기법의 적용 가능성이 낮거나 효과가 제한적일 수 있다.
%
% Second, 제안한 방어 기법(PAD 및 C-PAD)은 Transformer 구조의 self-attention 메커니즘을 기반으로 하므로, attention이 사용되지 않는 모델(CNN, RNN 기반 모델 등)에는 적용이 불가능하다. 따라서 실험 가능한 모델 구조에 제한이 있으며, 저사양 환경에서는 연산 비용이 부담이 될 수 있다.
%
% Third, 방어 기법의 성능은 외부 phonemizer의 품질과 언어 지원 범위에 크게 의존한다. 만약 phonemizer가 해당 언어를 지원하지 않거나 음소 추출 정확도가 낮은 경우, 방어 기법의 효과는 크게 감소할 수 있다. 이에 따라 저자원이거나 소수 언어에 대한 적용 가능성은 제한적 일 수 있다.
%
% Lastly, 욕설 및 혐오 표현은 끊임없이 변화하고 새롭게 생성되기 때문에, 정적인 데이터셋 기반 학습 방식에는 일반화의 한계가 존재한다. 특히, 우회 표현이나 신조어와 같은 새로운 공격 방식이 등장할 경우, 기존 학습된 모델이 이를 효과적으로 탐지하지 못할 가능성이 있다.
%
% ----
%
% 영작 부분:
First, PHISH may not be universally applicable across all languages.
Specifically, PHISH is designed under the assumption that human readers can easily infer the original text from its perturbed form.
As previously discussed, this assumption generally holds in languages with shallow orthographic depth, such as Korean, but may not hold in languages with deeper orthographic systems.

Second, seq-MESH and dir-MESH are inherently tied to transformer-based architectures that rely on attention mechanisms. This architectural dependence limits the applicability of our defense methods to models without self-attention, such as CNNs \cite{cnn} or traditional RNN-based classifiers. In addition, integrating phoneme-level information through additional cross-attention mechanism introduces computational overhead, which may hinder deployment in resource-constrained environments.

Lastly, the effectiveness of seq-MESH and dir-MESH requires an external phonemizer to generate phoneme sequences.
This means that the accuracy of such a phonemizer can affect the performance of our defense methods.
However, since we used the phonemizer without any optimization or refinement, we believe the reported performance represented in our paper could be improved by using a more accurate phonemizer.

\section*{Acknowledgments}
% 사사표기. AI대학원 100%.
This work was supported by the Institute of Information \& Communications Technology Planning \& Evaluation (IITP) grant funded by the Korea government (MSIT) [RS-2021-II211341, Artificial Intelligence Graduate School Program (Chung-Ang University)]
\bibliography{custom}

\begin{thebibliography}{24}
\providecommand{\natexlab}[1]{#1}

\bibitem[{Aggarwal and Zesch(2022)}]{global-visual}
Piush Aggarwal and Torsten Zesch. 2022.
\newblock \href {https://aclanthology.org/2022.wnut-1.25/} {Analyzing the real vulnerability of hate speech detection systems against targeted intentional noise}.
\newblock In \emph{Proceedings of the Eighth Workshop on Noisy User-generated Text (W-NUT 2022)}, pages 230--242, Gyeongju, Republic of Korea. Association for Computational Linguistics.

\bibitem[{Boucher et~al.(2022)Boucher, Shumailov, Anderson, and Papernot}]{badcharacter}
Nicholas Boucher, Ilia Shumailov, Ross Anderson, and Nicolas Papernot. 2022.
\newblock \href {https://doi.org/10.1109/SP46214.2022.9833641} {Bad characters: Imperceptible nlp attacks}.
\newblock In \emph{2022 IEEE Symposium on Security and Privacy (SP)}, pages 1987--2004.

\bibitem[{Cooper et~al.(2023)Cooper, Surdeanu, and Blanco}]{Homoglyphs}
Portia Cooper, Mihai Surdeanu, and Eduardo Blanco. 2023.
\newblock \href {https://doi.org/10.18653/v1/2023.findings-emnlp.192} {Hiding in plain sight: Tweets with hate speech masked by homoglyphs}.
\newblock In \emph{Findings of the Association for Computational Linguistics: EMNLP 2023}, pages 2922--2929, Singapore. Association for Computational Linguistics.

\bibitem[{El~Boukkouri et~al.(2020)El~Boukkouri, Ferret, Lavergne, Noji, Zweigenbaum, and Tsujii}]{character-bert}
Hicham El~Boukkouri, Olivier Ferret, Thomas Lavergne, Hiroshi Noji, Pierre Zweigenbaum, and Jun{'}ichi Tsujii. 2020.
\newblock \href {https://doi.org/10.18653/v1/2020.coling-main.609} {{C}haracter{BERT}: Reconciling {ELM}o and {BERT} for word-level open-vocabulary representations from characters}.
\newblock In \emph{Proceedings of the 28th International Conference on Computational Linguistics}, pages 6903--6915, Barcelona, Spain (Online). International Committee on Computational Linguistics.

\bibitem[{Ellis et~al.(2004)Ellis, Natsume, Stavropoulou, Hoxhallari, Van~Daal, Polyzoe, TSIPA, and Petalas}]{orthographic-depth}
Nick~C Ellis, Miwa Natsume, Katerina Stavropoulou, Lorenc Hoxhallari, Victor~HP Van~Daal, Nicoletta Polyzoe, MARIA-LOUISA TSIPA, and Michalis Petalas. 2004.
\newblock The effects of orthographic depth on learning to read alphabetic, syllabic, and logographic scripts.
\newblock \emph{Reading research quarterly}, 39(4):438--468.

\bibitem[{Jeong et~al.(2022)Jeong, Oh, Lee, Ahn, Moon, Park, and Oh}]{kold}
Younghoon Jeong, Juhyun Oh, Jongwon Lee, Jaimeen Ahn, Jihyung Moon, Sungjoon Park, and Alice Oh. 2022.
\newblock \href {https://doi.org/10.18653/v1/2022.emnlp-main.744} {{KOLD}: {K}orean offensive language dataset}.
\newblock In \emph{Proceedings of the 2022 Conference on Empirical Methods in Natural Language Processing}, pages 10818--10833, Abu Dhabi, United Arab Emirates. Association for Computational Linguistics.

\bibitem[{Kim(2011)}]{orthographic_feature}
Young-Suk Kim. 2011.
\newblock \href {https://doi.org/10.1016/j.cedpsych.2010.06.003} {Considering linguistic and orthographic features in early literacy acquisition: Evidence from korean}.
\newblock \emph{Contemporary Educational Psychology}, 36(3):177--189.

\bibitem[{Krizhevsky et~al.(2012)Krizhevsky, Sutskever, and Hinton}]{cnn}
Alex Krizhevsky, Ilya Sutskever, and Geoffrey~E Hinton. 2012.
\newblock \href {https://proceedings.neurips.cc/paper_files/paper/2012/file/c399862d3b9d6b76c8436e924a68c45b-Paper.pdf} {Imagenet classification with deep convolutional neural networks}.
\newblock In \emph{Advances in Neural Information Processing Systems}, volume~25. Curran Associates, Inc.

\bibitem[{Laboreiro and Oliveira(2014)}]{portuguese}
Gustavo Laboreiro and Eug{\'e}nio Oliveira. 2014.
\newblock What we can learn from looking at profanity.
\newblock In \emph{Computational Processing of the Portuguese Language: 11th International Conference, PROPOR 2014, S{\~a}o Carlos/SP, Brazil, October 6-8, 2014. Proceedings 11}, pages 108--113. Springer.

\bibitem[{Le et~al.(2022)Le, Lee, Yen, Hu, and Lee}]{perturbation_in_wild}
Thai Le, Jooyoung Lee, Kevin Yen, Yifan Hu, and Dongwon Lee. 2022.
\newblock \href {https://doi.org/10.18653/v1/2022.findings-acl.232} {Perturbations in the wild: Leveraging human-written text perturbations for realistic adversarial attack and defense}.
\newblock In \emph{Findings of the Association for Computational Linguistics: ACL 2022}, pages 2953--2965, Dublin, Ireland. Association for Computational Linguistics.

\bibitem[{Le et~al.(2023)Le, Ye, Hu, and Lee}]{cryptext}
Thai Le, Yiran Ye, Yifan Hu, and Dongwon Lee. 2023.
\newblock \href {https://doi.org/10.1109/ICDE55515.2023.00287} {Cryptext: Database and interactive toolkit of human-written text perturbations in the wild}.
\newblock In \emph{2023 IEEE 39th International Conference on Data Engineering (ICDE)}, pages 3639--3642.

\bibitem[{Lee et~al.(2025)Lee, Lee, Lee, and Lee}]{visual-abuse}
Hanyong Lee, Chaelyn Lee, Yongjae Lee, and Jaesung Lee. 2025.
\newblock \href {https://aclanthology.org/2025.findings-naacl.247/} {{B}it{A}buse: A dataset of visually perturbed texts for defending phishing attacks}.
\newblock In \emph{Findings of the Association for Computational Linguistics: NAACL 2025}, pages 4367--4384, Albuquerque, New Mexico. Association for Computational Linguistics.

\bibitem[{Lee(2020)}]{kcbert}
Junbum Lee. 2020.
\newblock Kcbert: Korean comments bert.
\newblock In \emph{Proceedings of the 32nd Annual Conference on Human and Cognitive Language Technology}, pages 437--440.

\bibitem[{Park et~al.(2023)Park, Kim, Park, and Park}]{khaters}
Chaewon Park, Suhwan Kim, Kyubyong Park, and Kunwoo Park. 2023.
\newblock K-haters: A hate speech detection corpus in korean with target-specific ratings.
\newblock \emph{Findings of the EMNLP 2023}.

\bibitem[{Park et~al.(2021{\natexlab{a}})Park, Kim, Cho, Park, Park, Kim, Chung, and Lee}]{koas}
San-Hee Park, Kang-Min Kim, Seonhee Cho, Jun-Hyung Park, Hyuntae Park, Hyuna Kim, Seongwon Chung, and SangKeun Lee. 2021{\natexlab{a}}.
\newblock \href {https://doi.org/10.18653/v1/2021.emnlp-demo.9} {{KOAS}: {K}orean text offensiveness analysis system}.
\newblock In \emph{Proceedings of the 2021 Conference on Empirical Methods in Natural Language Processing: System Demonstrations}, pages 72--78, Online and Punta Cana, Dominican Republic. Association for Computational Linguistics.

\bibitem[{Park et~al.(2021{\natexlab{b}})Park, Moon, Kim, Cho, Han, Park, Song, Kim, Song, Oh, Lee, Oh, Lyu, Jeong, Lee, Seo, Lee, Kim, Lee, Jang, Do, Kim, Lim, Lee, Park, Shin, Kim, Park, Park, Oh, Ha~(NAVER AI~Lab), Cho, and Cho}]{klue}
Sungjoon Park, Jihyung Moon, Sungdong Kim, Won~Ik Cho, Ji~Yoon Han, Jangwon Park, Chisung Song, Junseong Kim, Youngsook Song, Taehwan Oh, Joohong Lee, Juhyun Oh, Sungwon Lyu, Younghoon Jeong, Inkwon Lee, Sangwoo Seo, Dongjun Lee, Hyunwoo Kim, Myeonghwa Lee, Seongbo Jang, Seungwon Do, Sunkyoung Kim, Kyungtae Lim, Jongwon Lee, Kyumin Park, Jamin Shin, Seonghyun Kim, Lucy Park, Lucy Park, Alice Oh, Jung-Woo Ha~(NAVER AI~Lab), Kyunghyun Cho, and Kyunghyun Cho. 2021{\natexlab{b}}.
\newblock \href {https://datasets-benchmarks-proceedings.neurips.cc/paper_files/paper/2021/file/98dce83da57b0395e163467c9dae521b-Paper-round2.pdf} {Klue: Korean language understanding evaluation}.
\newblock In \emph{Proceedings of the Neural Information Processing Systems Track on Datasets and Benchmarks}, volume~1.

\bibitem[{Perea and Lupker(2004)}]{transposedletter}
Manuel Perea and Stephen~J Lupker. 2004.
\newblock Can caniso activate casino? transposed-letter similarity effects with nonadjacent letter positions.
\newblock \emph{Journal of memory and language}, 51(2):231--246.

\bibitem[{Puertas and Martinez-Santos(2021)}]{Phoneticdetection}
Edwin Puertas and Juan~Carlos Martinez-Santos. 2021.
\newblock Phonetic detection for hate speech spreaders on twitter.

\bibitem[{Rothe et~al.(2020)Rothe, Narayan, and Severyn}]{bertgeneration}
Sascha Rothe, Shashi Narayan, and Aliaksei Severyn. 2020.
\newblock \href {https://doi.org/10.1162/tacl_a_00313} {Leveraging pre-trained checkpoints for sequence generation tasks}.
\newblock \emph{Transactions of the Association for Computational Linguistics}, 8:264--280.

\bibitem[{Seth et~al.(2023)Seth, Stureborg, Pruthi, and Dhingra}]{visualperturbations}
Dev Seth, Rickard Stureborg, Danish Pruthi, and Bhuwan Dhingra. 2023.
\newblock \href {https://doi.org/10.18653/v1/2023.eacl-main.238} {Learning the legibility of visual text perturbations}.
\newblock In \emph{Proceedings of the 17th Conference of the European Chapter of the Association for Computational Linguistics}, pages 3260--3273, Dubrovnik, Croatia. Association for Computational Linguistics.

\bibitem[{Shekhar and Venkatesan(2018)}]{bagofphonetic}
Ankita Shekhar and M.~Venkatesan. 2018.
\newblock \href {https://doi.org/10.1109/ICCTCT.2018.8550938} {A bag-of-phonetic-codes modelfor cyber-bullying detection in twitter}.
\newblock In \emph{2018 International Conference on Current Trends towards Converging Technologies (ICCTCT)}, pages 1--7.

\bibitem[{Yang and Lin(2020)}]{tocp}
Hsu Yang and Chuan-Jie Lin. 2020.
\newblock \href {https://aclanthology.org/2020.trac-1.2/} {{TOCP}: A dataset for {C}hinese profanity processing}.
\newblock In \emph{Proceedings of the Second Workshop on Trolling, Aggression and Cyberbullying}, pages 6--12, Marseille, France. European Language Resources Association (ELRA).

\bibitem[{Yi et~al.(2021)Yi, Lim, Ko, and Shin}]{korean-embeddinglstm}
MoungHo Yi, MyungJin Lim, Hoon Ko, and JuHyun Shin. 2021.
\newblock Method of profanity detection using word embedding and lstm.
\newblock \emph{Mobile Information Systems}, 2021(1):6654029.

\bibitem[{Yu et~al.(2024)Yu, Choi, and Kim}]{koreanfool}
Seunguk Yu, Juhwan Choi, and YoungBin Kim. 2024.
\newblock \href {https://doi.org/10.18653/v1/2024.findings-naacl.219} {Don`t be a fool: Pooling strategies in offensive language detection from user-intended adversarial attacks}.
\newblock In \emph{Findings of the Association for Computational Linguistics: NAACL 2024}, pages 3456--3467, Mexico City, Mexico. Association for Computational Linguistics.

\end{thebibliography}
% \section*{Acknowledgments}
% 사사표기. AI대학원 100%.
\appendix

% \section{Additioanl Experimental Result}
% \label{app:additional-result}
% To propose more detailed experimental results, we conducted additional experiments by attacking the test sets of the K-HATERS and KoLD datasets with 5\%, 15\%, and 25\% perturbation ratios, which were not covered in Section \ref{sec:result}.

% Table \ref{tab:khaters-single-result}, \ref{tab:khaters-double-result}, \ref{tab:kold-single-result}, and \ref{tab:kold-double-result} shows the 

% \input{table/khaters_single_result}
% \input{table/khaters_double_result}
% \input{table/kold_single_result}
% \input{table/kold_double_result}

% \section{Korean Jamos}
% \input{table/vowel_ipa}
%\input{table/consonant_ipa}

% \section{Experimental Result with Dual-Jamo Attack}
% \input{table/khaters_double_full}
% \input{table/kold_double_full}
% Table \ref{tab:khaters-double-result-full} on page \pageref{tab:khaters-double-result-full} and \ref{tab:kold-double-result-full} on page \pageref{tab:kold-double-result-full} shows the experimental result using dual-jamo attack.

\section{Look-up table}
\label{app:lookup}
% Please add the following required packages to your document preamble:
% \usepackage{multirow}
\begin{table}[t]
\centering
\begin{tabular}{lcp{0.55\columnwidth}}
\toprule
Type                               & Base & Jamo set                                \\
\midrule
\multirow{5}{*}{Onset} & \textipa{/k/}         & \{ㄱ, ㄲ, ㅋ\}                     \\
                                   & \textipa{/t/}         & \{ㄷ, ㄸ, ㅌ\}                     \\
                                   & \textipa{/p/}         & \{ㅂ, ㅃ, ㅍ\}                     \\
                                   & \textipa{/t\textctc/}         & \{ㅈ, ㅉ, ㅊ\}                     \\
                                   & \textipa{/s/}         & \{ㅅ, ㅆ \}                         \\
                                   \midrule
\multirow{7}{*}{Nucleus}             & \textipa{/i/}         & \{ㅣ, ㅢ\}                          \\
                                   & \textipa{/u/}         & \{ㅜ, ㅠ\}                          \\
                                   & \textipa{/o/}         & \{ㅗ, ㅛ\}                          \\
                                   & \textipa{/2/}         & \{ㅓ, ㅝ, ㅕ\}                     \\
                                   & \textipa{/a/}         & \{ㅏ, ㅘ, ㅑ\}                     \\
                                   & \textipa{/e/}        & \{ㅔ, ㅞ, ㅖ\}                     \\
                                   & \textipa{/E/}         & \{ㅐ, ㅙ, ㅒ\}                     \\
                                   \midrule
\multirow{6}{*}{Coda}   &    \textipa{/k/}      & \{ㄱ, ㄲ, ㅋ, ㄳ, ㄺ\}                           \\
                                 &      \textipa{/n/}    & \{ㄴ,ㄵ, ㄶ  \}                   \\
                                   &     \textipa{/t/}    & \{ㅅ, ㅆ, ㄷ, ㅌ, ㅈ, ㅊ, ㅎ\} \\
                                   &      \textipa{/l/}    & \{ㄹ,ㄺ, ㄼ, ㄽ, ㄾ, ㅀ\}    \\
                                   &      \textipa{/m/}    & \{ㅁ,ㄻ\}                        \\
                                   &      \textipa{/p/}    & \{ㅂ,ㅍ, ㄼ, ㅄ, ㄿ\} \\         

\bottomrule
\end{tabular}%
% }
\caption{Predefined look-up table}
\label{tab:lookup}
\end{table}
Table \ref{tab:lookup} illustrates the predefined look-up table for Korean initial consonants (onset), vowels (nucleus), and final consonants (coda). Jamos assigned to the same set can be substituted with others in the same set.
Each IPA symbol of the initial consonants (onset) and vowels (nucleus) indicates the base phone of the corresponding jamo set.
Additionally, final consonants (coda) are pronounced as their corresponding base phones according to the Korean standard pronunciation rule.

\section{Statistics of Texts}
\label{app:statistics}
Tables \ref{tab:single_unk} and \ref{tab:dual_unk} on page \pageref{tab:dual_unk} present the appearance rates of unknown tokens in both text and phoneme sequences across different detectors after conducting our PHISH attack.
In both tables, BERT and RoBERTa show the same statistics since they were pretrained on the same corpus.
Notably, KCBERT exhibits a lower rate of unknown tokens in the text than the other two detectors. This gap remains relatively small even when the input text is perturbed.
We speculate that this robustness stems from KCBERT's pretraining data, which includes comments posted on online news articles, potentially containing naturally perturbed texts authored by malicious users.

These statistics also offer additional insights into our experimental results.
First, the statistics can explain why augmenting phoneme sequences helps mitigate semantic loss caused by phonetic perturbations. When we use a higher attack ratio, the number of unknown tokens increases. So, current models may suffer semantic loss or distortion due to PHISH's phonetic perturbations. By providing phonetic information to the detectors, we could mitigate this loss.

Second, the statistics may explain why KCBERT consistently outperforms the other two detectors. As KCBERT showed fewer unknown tokens, it is highly likely that the model suffers less from semantic loss than the other two models. So, it could achieve higher performance by incorporating semantic and phonetic information, without a considerable loss.

\begin{table}
\centering
\small
% \resizebox{\textwidth}{!}{%
% Please add the following required packages to your document preamble:
% \usepackage{graphicx}
% \usepackage[table,xcdraw]{xcolor}
% Beamer presentation requires \usepackage{colortbl} instead of \usepackage[table,xcdraw]{xcolor}
% Please add the following required packages to your document preamble:
% \usepackage{graphicx}
% \usepackage[table,xcdraw]{xcolor}
% Beamer presentation requires \usepackage{colortbl} instead of \usepackage[table,xcdraw]{xcolor}
% Please add the following required packages to your document preamble:
% \usepackage{graphicx}
% \usepackage[table,xcdraw]{xcolor}
% Beamer presentation requires \usepackage{colortbl} instead of \usepackage[table,xcdraw]{xcolor}
% \begin{tabular}{lr|r@{$\pm$}lr@{$\pm$}lr@{$\pm$}lr@{$\pm$}lr@{$\pm$}lr@{$\pm$}l}
\begin{tabular}{@{}ll@{\;}c@{\;}r@{$\pm$}rr@{$\pm$}r@{}}
\toprule
Model            & Dataset          & \multicolumn{1}{l}{Attack}     & \multicolumn{2}{c}{Text}   & \multicolumn{2}{c}{Phoneme}\\ 
 &&Ratio(\%)  &\multicolumn{2}{c}{UNK avg}  &\multicolumn{2}{c}{UNK avg}\\
\midrule
BERT            & K-HATERS      & 0              & 0.4 & 1.9      &3.5&5.4   \\
&                &               10             & 5.3 & 5.6         &5.3&6.3\\
&                &               20             & 11.8 & 9.3           &7.5&7.7         \\
&                &               30             & 17.9 & 12.4          &9.8&9.9      \\ 
\cmidrule{2-7}
&                KoLD          & 0              & 0.6 & 4.0          &4.0&7.8  \\
&                &               10             & 5.6 & 7.4          &5.7&8.5      \\ 
&                &               20             & 13.0 & 12.5          &8.4&11.1     \\ 
&                &                30             & 19.2 & 15.5          &10.4&12.5     \\

\midrule
RoBERTa            & K-HATERS      & 0          & 0.4 & 1.9         &3.5&5.4   \\
&                &               10             & 5.3 & 5.6         &5.3&6.3 \\
&                &               20             & 11.8 & 9.3         &7.5&7.7  \\
&                &               30             & 17.9 & 12.4        &9.8&9.9     \\ 
\cmidrule{2-7}
&                KoLD          & 0              & 0.6 & 4.0        &4.0&7.8    \\
&                &               10             & 5.6 & 7.4         &5.7&8.5        \\ 
&                &               20             & 13.0 & 12.5         &8.4&11.1      \\ 
&                &                30            & 19.2 & 15.5            &10.4&12.5    \\

\midrule
KCBERT            & K-HATERS      & 0           & 0.5 & 3.3      &1.1&3.5  \\
&                &               10             & 1.9 & 4.4       &1.2&3.5   \\
&                &               20             & 3.4 & 5.8            &1.4&3.6      \\
&                &               30             & 4.7 & 6.5         &1.6&4.0      \\ 
\cmidrule{2-7}
&                KoLD          & 0              & 0.5 & 3.7         &1.0&4.3 \\
&                &               10             & 1.8 & 5.2          &1.1&4.4     \\ 
&                &               20             & 3.4 & 6.9          &1.4&4.7     \\ 
&                &                30            & 4.7 & 8.7           &1.5&4.9      \\

\bottomrule

\end{tabular}%
% }
\caption{Statistics of unknown tokens in perturbed texts using single-jamo attack and their phoneme sequences}
\label{tab:single_unk}
\end{table}
\begin{table} %[!htp]
\centering
\small
% \resizebox{\textwidth}{!}{%
% Please add the following required packages to your document preamble:
% \usepackage{graphicx}
% \usepackage[table,xcdraw]{xcolor}
% Beamer presentation requires \usepackage{colortbl} instead of \usepackage[table,xcdraw]{xcolor}
% Please add the following required packages to your document preamble:
% \usepackage{graphicx}
% \usepackage[table,xcdraw]{xcolor}
% Beamer presentation requires \usepackage{colortbl} instead of \usepackage[table,xcdraw]{xcolor}
% Please add the following required packages to your document preamble:
% \usepackage{graphicx}
% \usepackage[table,xcdraw]{xcolor}
% Beamer presentation requires \usepackage{colortbl} instead of \usepackage[table,xcdraw]{xcolor}
% \begin{tabular}{lr|r@{$\pm$}lr@{$\pm$}lr@{$\pm$}lr@{$\pm$}lr@{$\pm$}lr@{$\pm$}l}
\begin{tabular}{@{}ll@{\;}c@{\;}r@{$\pm$}rr@{$\pm$}r@{}}
\toprule
Model            & Dataset          & \multicolumn{1}{l}{Attack}     & \multicolumn{2}{c}{Text}   & \multicolumn{2}{c}{Phoneme}\\ 
 &&Ratio(\%)  &\multicolumn{2}{c}{UNK avg}  &\multicolumn{2}{c}{UNK avg}\\
%&&Ratio(\%)\\
\midrule
BERT            & K-HATERS      & 0              & 0.4  & 1.9       &3.5&5.4      \\
&                &               10             & 9.9 & 6.9           &8.1&7.2   \\
&                &               20             & 22.9 & 11.7        &14.7&10.8     \\
&                &               30             & 34.2 & 15.7        &20.7&13.4     \\ 
\cmidrule{2-7}
&                KoLD          & 0              & 0.6 & 4.0          &4.0&7.8    \\
&                &               10             & 10.0 & 8.2         &8.6&9.7     \\ 
&                &               20             & 24.8 & 14.8        &15.9&13.8     \\ 
&                &                30             & 37.1 & 18.3        &22.5&16.9     \\

\midrule
RoBERTa            & K-HATERS      & 0          & 0.4 & 1.9        &3.5&5.4    \\
&                &               10             & 9.9 & 6.9        &8.1&7.2    \\
&                &               20             & 22.9 & 11.7      &14.7&10.8     \\
&                &               30             & 34.2 & 15.7      &20.7&13.4       \\ 
\cmidrule{2-7}
&                KoLD          & 0              & 0.6 & 4.0         &4.0&7.8      \\
&                &               10             & 10.0 & 8.2         &8.6&9.7     \\ 
&                &               20             & 24.8 & 14.8        &15.9&13.8    \\ 
&                &                30            & 37.1 & 18.3         &22.5&16.9    \\

\midrule
KCBERT            & K-HATERS      & 0           & 0.5 & 3.3        &1.1&3.5     \\
&                &               10             & 6.2 & 7.0         &1.8&4.0   \\
&                &               20             & 12.7 & 9.9        &2.7&5.1     \\
&                &               30             & 18.5 & 12.7        &3.5&5.6    \\ 
\cmidrule{2-7}
&                KoLD          & 0              & 0.5 & 3.7          &1.0&4.3    \\
&                &               10             & 6.3 & 7.9         &1.9&5.7     \\ 
&                &               20             & 14.1 & 13.3         &3.0&7.3    \\ 
&                &                30            & 20.2 & 15.8         &4.0&8.6     \\

\bottomrule

\end{tabular}%
% }
\caption{Statistics of unknown tokens in perturbed texts using dual-jamo attack and their phoneme sequences}
\label{tab:dual_unk}
\end{table}

% As shown in Table \ref{tab:single_unk} on page \pageref{tab:single_unk}, the proportion of unknown tokens in the perturbed text rises significantly with higher attack ratios.
% \isitnecessary
% As the training data of KCBERT contains online comments, KCBERT가 상대적으로 더 perturbed된 text를 training 과정에서 마주했을 가능성이 높고 이로 인해 unknown token의 비율이 낮아 결론적으로 robustness가 더 높은 것으로 해석한다.

\end{document}